%% file: main.tex
\documentclass[conference]{IEEEtran}
\IEEEoverridecommandlockouts

\usepackage[utf8]{inputenc} 
\usepackage[T1]{fontenc}

\input{math_commands.tex}

\usepackage[style=numeric]{biblatex}
\usepackage{hyperref}

\usepackage{amsmath,amssymb,amsfonts}
\usepackage{graphicx}
\usepackage{textcomp}
\usepackage{xcolor}
\def\BibTeX{{\rm B\kern-.05em{\sc i\kern-.025em b}\kern-.08em
    T\kern-.1667em\lower.7ex\hbox{E}\kern-.125emX}}

\usepackage{tocbibind}
\usepackage{diagbox}
\usepackage{paralist}
\usepackage{dsfont}
\usepackage{diagbox}
\usepackage{enumitem}
\usepackage{xspace}
\usepackage[linesnumbered,ruled,vlined]{algorithm2e}

\newcommand{\him}{HIGhER\xspace}
\newcommand{\himlong}{Hindsight Generation for Experience Replay\xspace}

\newcommand{\chapternote}[1]{{%
  \let\thempfn\relax
  \footnotetext[0]{\emph{#1}}
}}

\begin{document}
\title{HIGhER : Improving instruction following with \\Hindsight Generation for Experience Replay}
\author{
\IEEEauthorblockN{Geoffrey Cideron*}
\IEEEauthorblockA{\textit{Université de Lille} \\ CRIStAL, CNRS, Inria \\
France}
\and
\IEEEauthorblockN{Mathieu Seurin*}
\IEEEauthorblockA{\textit{Université de Lille}\\ CRIStAL, CNRS, Inria\\
France}
\and
\IEEEauthorblockN{Florian Strub}
\IEEEauthorblockA{\textit{DeepMind} \\ Paris\\
France}
\and
\IEEEauthorblockN{Olivier Pietquin}
\IEEEauthorblockA{\textit{Google Research} \\ Brain Team, Paris\\
France}
}


\IEEEoverridecommandlockouts
\IEEEpubid{\makebox[\columnwidth]
{978-1-7281-2547-3/20/\$31.00~\copyright2020 IEEE \hfill} 
\hspace{\columnsep}\makebox[\columnwidth]{ }}


\maketitle
\IEEEpubidadjcol

\begin{abstract}
Language creates a compact representation of the world and allows the description of unlimited situations and objectives through compositionality. While these characterizations may foster instructing, conditioning or structuring interactive agent behavior, it remains an open-problem to correctly relate language understanding and reinforcement learning in even simple instruction following scenarios. This joint learning problem is alleviated through expert demonstrations, auxiliary losses, or neural inductive biases. In this paper, we propose an orthogonal approach called \himlong (\him) that extends the Hindsight Experience Replay approach to the language-conditioned policy setting. Whenever the agent does not fulfill its instruction, \him learns to output a new directive that matches the agent trajectory, and it relabels the episode with a positive reward. To do so, \him learns to map a state into an instruction by using past successful trajectories, which removes the need to have external expert interventions to relabel episodes as in vanilla HER. We show the efficiency of our approach in the BabyAI environment, and demonstrate how it complements other instruction following methods.
\end{abstract}


\section{Introduction}

\chapternote{* Those authors contributed equally}

Language has slowly evolved to communicate intents, to state objectives, or to describe complex situations~\cite{kirby2015compression}. It conveys information compactly by relying on composition and highlighting salient facts.  As language can express a vast diversity of goals and situations, it may help conditioning the training of interactive agents over heterogeneous and composite tasks~\cite{luketina2019survey} and help transfer \cite{narasimhan2018grounding}.
Unfortunately, conditioning a policy on language also entails a supplementary difficulty as the agent needs to understand linguistic cues to alter its behavior. The agent thus needs to ground its language understanding by relating the words to its observations, actions, and rewards before being able to leverage the language structure~\cite{kiela2016virtual,hermann2017grounded}. Once the linguistic symbols are grounded, the agent may then take advantage of language compositionality to condition its policy on new goals. 

In this work, we use instruction following as a natural testbed to examine this question~\cite{tellex2011understanding,chen2011learning,artzi2013weakly,luketina2019survey, hermann2019learning}. In this setting, the agent is given a text description of its goal (e.g. "pick the red ball") and is rewarded when achieving it.
The agent has thus to visually ground the language, i.e., linking and disentangling visual attributes (\emph{shape}, \emph{color}) from language description ("ball", "red") by using rewards to condition its policy toward task completion. On one side, the language compositionality allows for a high number of goals, and offers generalization opportunities; but on the other side, it dramatically complexifies the policy search space. Besides, instruction following is a notoriously hard RL problem as the training signal is very sparse since the agent is only rewarded over task completion.
In practice, the navigation and language grounding problems are often circumvented by warm-starting the policy with labeled trajectories~\cite{zang2018translating,anderson2018vision}.
Although scalable, these approaches require numerous human demonstrations, whereas we here want to jointly learn the navigation policy and language understanding from scratch. In a seminal work, \cite{hermann2017grounded} successfully ground language instructions, but the authors used unsupervised losses and heavy curriculum to handle the sparse reward challenge. 

\begin{figure*}
         \centering
         \includegraphics[width=\textwidth]{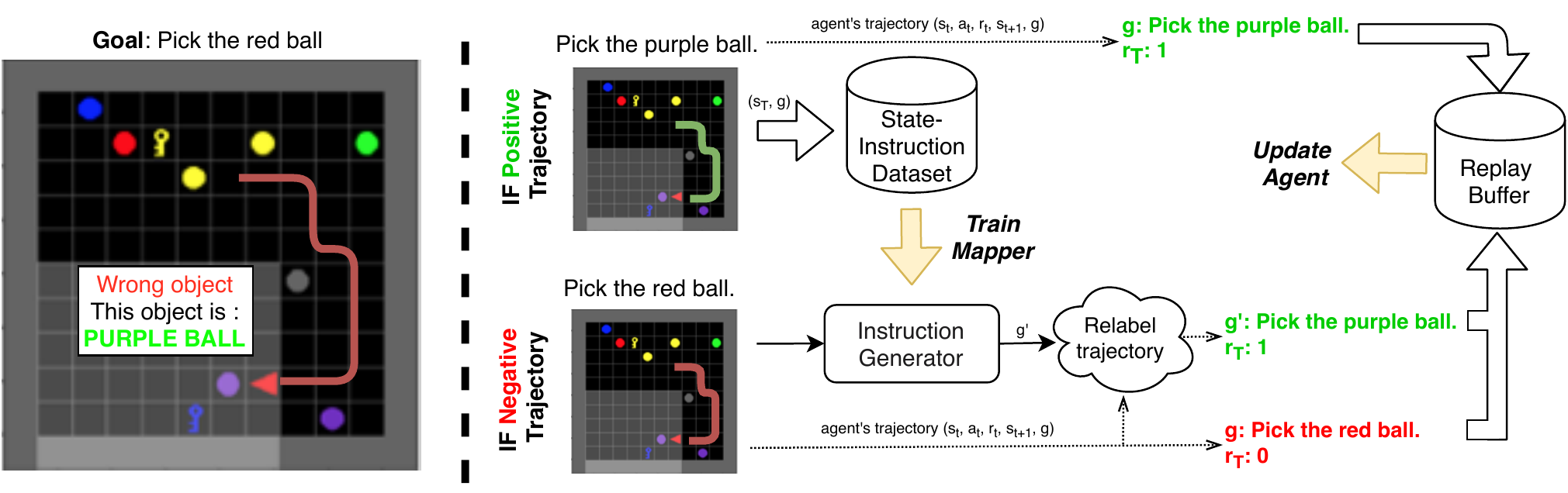}
        \caption{Upon positive trajectory, the agent trajectory is added to the RL replay buffer and the goal mapper dataset. Upon failed trajectory, the goal mapper is used to relabel the
        episode, and both trajectories are appended to the replay buffer. In the original HER paper, the mapping function is bypassed since they are dealing with spatial goals, and therefore, vanilla HER cannot be applied without external expert.}
        \label{fig:sketch}
\vskip -1em
\end{figure*}

In this paper, we take advantage of language compositionality to tackle the lack of reward signals. To do so, we extend Hindsight Experience Replay (HER) to language goals~\cite{andrychowicz2017hindsight}. HER originally deals with the sparse reward problems in spatial scenario;  it relabels unsuccessful trajectories into successful ones by redefining the policy goal \emph{a posteriori}. As a result, HER creates additional episodes with positive rewards and a more diverse set of goals.
Unfortunately, this approach cannot be directly applied when dealing with linguistic goals. As HER requires a mapping between the agent trajectory and the goal to substitute, it requires expert supervision to describe failed episodes with words. Hence, this mapping should either be handcrafted with synthetic bots~\cite{chan2019actrce}, or be learned from human demonstrations, which would both limit HER generality. More generally, language adds a level of semantics, which allows generating textual objective that could not be encoded by simple spatial observations as in regular HER, e.g., "fetch a ball that is not blue" or "pick any red object".

%
In this work, we introduce \himlong (\him), a training procedure where the agent jointly learns the language-goal mapping and the navigation policy by solely interacting with the environment illustrated in \autoref{fig:sketch}. \him leverages positive trajectories to learn a mapping function, and \him then tackles the sparse reward problem by relabeling language goals upon negative trajectories in a HER fashion.  We evaluate our method on the BabyAI world~\cite{babyai_iclr19}, showing a clear improvement over RL baselines while highlighting the robustness of \him to noise. 

\section{Background And Notation}\label{sec:background}
In reinforcement learning, an agent interacts with the environment to maximize its cumulative reward~\cite{sutton2018reinforcement}. At each time step $t$,  
the agent is in a state $s_{t} \in \gS$, where it selects an action $a_t \in \gA$ according its policy $\pi: \gS \rightarrow \gA$. 
It then receives a reward $r_{t}$ from the environment's reward function 
$r: \gS \times \gA \rightarrow \mathds{R}$ and moves to the next state $s_{t+1}$ with probability
$p(s_{t+1}|s_{t},a_{t})$. 
The quality of the policy is assessed by the Q-function defined by  $Q^{\pi}(s,a) = \mathds{E}_{\pi} \left[ \sum_{t} \gamma^{t} r(s_{t},a_{t}) |  s_{0}=s, a_{0}=a\right]$ 
for all $(s,a)$ where $\gamma \in [0,1]$ is the discount factor. We define the optimal Q-value as $Q^{*}(s,a) = \max_{\pi}Q^{\pi}(s,a)$, from which the optimal policy $\pi^{*}$ is derived. 
%
We here use Deep Q-learning (DQN) to evaluate the optimal Q-function with neural networks and perform off-policy updates by sampling transitions $(s_{t}, a_{t}, r_{t}, s_{t+1})$ from a replay buffer~\cite{mnih2015human}. 

In this article, we augment our environment with a goal space $\gG$ which defines a new reward function $r: \gS \times \gA \times \gG \rightarrow \mathds{R}$ and policy $\pi: \gS \times \gG \rightarrow \gA$ by conditioning them on a goal descriptor $g \in \gG$. Similarly, the Q-function is also conditioned on the goal, and it is referred to as Universal Value Function Approximator (UVFA)~\cite{schaul2015universal}. This approach allows learning holistic policies that generalize over goals in addition to states at the expense of complexifying the training process. In this paper, we explore how language can be used for structuring the goal space, and how language composition eases generalization over unseen scenarios in a UVFA setting.

\paragraph{Hindsight Experience Replay (HER)} \cite{andrychowicz2017hindsight} is designed to increase the sample efficiency of off-policy RL algorithms such as DQN in the goal-conditioning setting. 
It reduces the sparse reward problem by taking advantage of failed trajectories, relabelling them with new goals. An expert then assigns the goal that was achieved by the agent when performing its trajectory, before updating the agent memory replay buffer with an additional positive trajectory.

Formally, HER assumes the existence of a predicate $f: \gS \times \gG \rightarrow \{0, 1\}$ which encodes whether the agent in a state $s$ satisfies the goal $f(s,g)=1$, and defines the  reward function $r(s_t,a,g)=f(s_{t+1},g)$. 
At the beginning of an episode, a goal $g$ is drawn from the space $\gG$ of goals. At each time step $t$, the transition $(s_{t}, a_{t}, r_{t}, s_{t+1}, g)$ is stored in the DQN replay buffer, and at the end of an unsuccessful episode, an expert provides an additional goal $g'$ that matches the trajectory. New transitions $(s_{t}, a_{t}, r'_{t}, s_{t+1}, g')$ are thus added to the replay buffer for each time step $t$, where $r'= r(s_{t}, a_{t}, s_{t+1}, g')$. DQN update rule remains identical to \cite{mnih2015human}, transitions are sampled from the replay buffer, and the network is updated using one step td-error minimization.


HER assumes that a mapping $m$ between states $s$ and goals $g$ is given.
In the original paper, this requirement is not restrictive as the goal space is a subset of the state space. Thus, the mapping $m$ is straightforward since any state along the trajectory can be used as a substitution goal. 
In the general case, the goal space differs from the state space, and the mapping function is generally unknown. In the instruction following setting, there is no obvious mapping from visual states to linguistic instructions. It thus requires expert intervention to provide a new language goal given the trajectory, which drastically reduces the interest of HER. Therefore, we here explore how to learn this mapping without any form of expert knowledge nor supervision.

\section{\himlong}\label{sec:him}


\himlong (\him) aims to learn a mapping from past experiences that relates a trajectory to a goal in order to apply HER, even when no expert are available. The mapping function relabels unsuccessful trajectories by predicting a substitute goal $\hat{g}$ as an expert would do. The transitions are then  appended to the replay buffer. This mapping learning is performed alongside agent policy training.

Besides, we wish to discard any form of expert supervision to learn this mapping as it would reduce the practicability of the approach. Therefore, the core idea is to use environment signals to retrieve training mapping pairs. Instinctively, in the sparse reward setting, trajectories with positive rewards encode ground-truth mapping pairs, while trajectories with negative rewards are mismatched pairs. These cues are thus collected to train the mapping function for \him in a supervised fashion. 
We emphasize that such signals are inherent to the environment, and an external expert does not provide them. In the following, we only keep positive pairs in order to train a discriminative mapping model.

Formally, \him is composed of a dataset $D$ of $\langle s, g \rangle$ pairs, a replay buffer $R$ and a parametrized mapping model $m_\vw$. For each episode, a goal $g$ is picked, and the agent generates transitions $(s_{t}, a_{t}, r_{t}, s_{t+1}, g)$ that are appended to the replay buffer $R$. The Q-function parameters are updated with an off-policy algorithm by sampling minibatches from $D$. Upon episode termination, if the goal is achieved, i.e. $f(s_{T},g)=1$, the $\langle s_{T}, g \rangle$ pair is appended to the dataset $D$. If the goal is not achieved, a substitute goal is sampled from the mapping model\footnote{The mapping model can be utilized with an accuracy criterion over a validation set to avoid random goal sampling.} $m_{\vw}(s_{T}) = \hat{g}'$ and the additional transitions $\{(s_{t}, a_{t}, r_{t}, s_{t+1}, \hat{g}')\}^T_{t=0}$ are added to the replay buffer. 
At regular intervals, the mapping model $m_\vw$ is optimized to predict the goal $g$ given the trajectory $\tau$ by sampling mini-batches from $D$.  Noticeably, \him can be extended to partially observable environments by replacing the predicate function $f(s,g)$ by $f(\tau,g)$, i.e., the completion of a goal depends on the full trajectory rather than one state. Although we assess \him in the instruction following setting, the proposed procedure can be extended to any other goal modalities.

\section{Experiments}


\subsection{Experimental Setting}\label{subsec:minigrid}

\paragraph{Environment} We experiment our approach on a visual domain called Minigrid \cite{babyai_iclr19}. This environment offers a variety of instruction-following tasks using a synthetic language for grounded language learning.
We use a 10x10 grid with 10 objects randomly located in the room. Each object has 4 attributes (shade, size, color, and type) inducing a total of 300 different objects (240 objects are used for training, 60 for testing). To the best of our knowledge, the number of different objects and its diversity is greater than concurrent works (\cite{chaplot2018gated} used 55 train instructions and 15 test instructions and \cite{hill2017understanding} has a total of 40 different objects).
The agent has four actions \{forward, left, right, pick\}, and it can only see the 7x7 grid in front of it.
For each episode, one object's attribute is randomly picked as a goal, and the text generator translates it in synthetic language.
, e.g., "Fetch a tiny light blue ball." The agent is rewarded when picking one object matching the goal description, which ends the episode; otherwise, the episode stops after 40 steps or after taking an incorrect object.

\paragraph{Task Complexity} It is important to underline the complexity of this task. To get rewards over multiple episodes, the agent must learn to navigate and inspect objects in the room while simultaneously learning the meaning of each word and how they relate to visual characteristics. The burden comes from reward sparsity as the replay buffer is filled with unsuccessful trajectories RL fails to learn. Alleviating this problem is essential and minigrid is an excellent testbed to assess algorithmic performances as an agent deals with partial observability, visual representation learning, and language grounding only from sparse rewards signal. 

\paragraph{Models}

In this experiment, \him is composed of two separate models. The instruction generator is a neural network outputting a sequence of words given the final state of a trajectory. 
It is trained by gradient descent using a cross-entropy loss on the dataset $D$ collected as described in \autoref{sec:him}. We train a DQN network following \cite{mnih2015human} with a dueling head~\cite{wang2016dueling}, double Q-learning ~\cite{hasselt2016deep}, and a prioritized replay buffer~\cite{schaul2015prioritized} over trajectories. The network receives a tuple $<s,g>$ as input and output an action corresponding to the argmax over states-actions values $Q(s,a,g)$. We use $\epsilon$-greedy exploration with decaying $\epsilon$.

\begin{algorithm}
    \small
    \SetAlgoLined
    \SetKwInput{KwInput}{Given}
    \KwInput{\begin{itemize}
    \item an off-policy RL algorithm (e.g. DQN) $\sA$
    \item a reward function $r: \gS \times \gA \times \gG \rightarrow \R$.
    \item a language score (e.g. parser accuracy, BLEU etc.) 
    \end{itemize}}
    Initialize $\sA$\ , replay buffer $R$, dataset $D_{train}$ and $D_{val}$ of $\langle instruction, state \rangle$, Instruction Generator $m_{\vw}$\;
    \For{episode=1,M}{
    Sample a goal $g$ and an initial state $s_{0}$\;
    $t=0$\;
    \Repeat{episode ends}{Execute an action $a_{t}$ chosen from the behavioral policy $\sA$: $a_{t} \leftarrow \pi(s_{t}||g)$\;
    Observe a reward $r_{t}=r(s_{t}, a_{t}, g)$ and a new state $s_{t+1}$\;
    Store the transition $(s_{t}, a_{t}, r_{t}, s_{t+1},g)$ in $R$\;
    Update Q-network parameters using the policy $\sA$ and sampled minibatches from $R$\;
    $t=t+1$\;}
    \If{$f(s_{t},g)=1$}{
    Store the pair $\langle s_{t},g \rangle$ in $D_{train}$ or $D_{val}$\;
    Update $m_w$ parameters by sampling minibatches from  $D_{train}$\;}
    \Else{
        \If{$m_w$ language validation score is high enough and $D_{val}$ is big enough}{
            Sample $\hat{g}'=m_{\vw}(s_{t})$\;
            Replace $g$ by $\hat{g}'$ in the transitions of the last episode and set $\hat{r}=r(s_{t}, a_{t}, \hat{g}')$.
        }
    }
    }
    \caption{\himlong (\him)}
    \label{algo:him}
\end{algorithm}

\subsection{Building Intuition}

This section examines the feasibility of \him by analysing two potential issues. We first show that HER is robust to a noisy mapping function (or partially incorrect goals), we then estimate the accuracy and generalisation performance of the instruction generator.  



\subsubsection{Noisy instruction generator and HER}\label{subsec:noisy_her}

\begin{figure}[t]

 \begin{minipage}{.45\textwidth}
 \centering
 \includegraphics[width=1\linewidth]{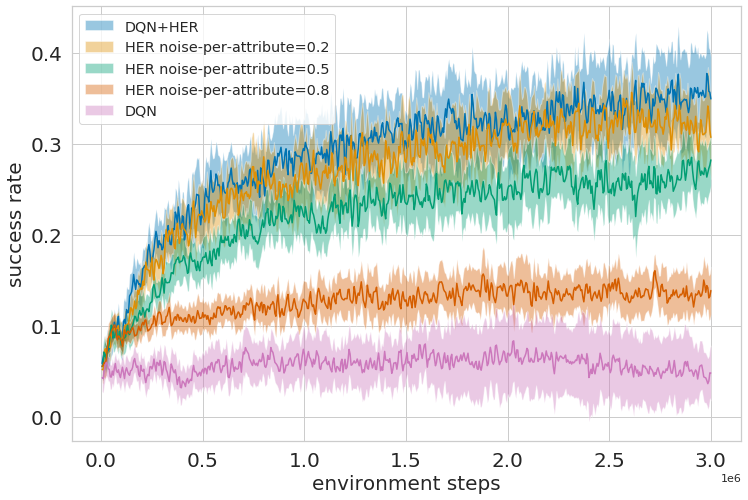}
 \end{minipage}
 \hfill
 \begin{minipage}{.45\textwidth}
 \centering
 \includegraphics[width=1\linewidth]{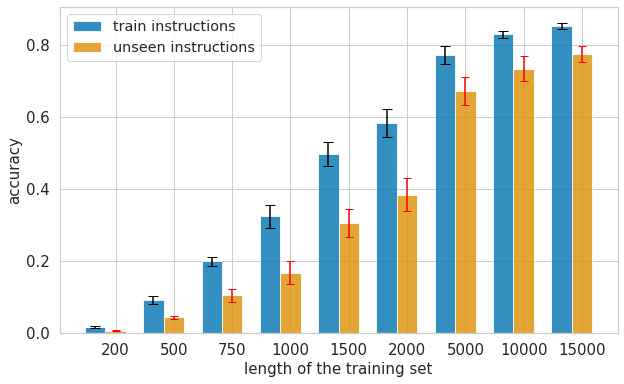}
 \label{fig:ig_merge}
 \end{minipage}

  \caption{\textbf{Top}: Agent performance with noisy mapping function. \textbf{Bottom}: Instruction generator accuracy over 5k pairs. Figures are averaged over 5 seeds and error bars shows one standard deviation.} 
  \label{fig:noisyher}
 \end{figure}

 We investigate how a noisy mapping $m$ affects performance compared to a perfect mapping. As the learned instruction generator is likely to be imperfect, it is crucial to assess how a noisy mapping may alter the training of the agent. To do so, we train an agent with HER and a synthetic bot to relabel unsuccessful trajectories. We then inject noise in our mapping where each attribute has a fixed probability \textit{p} to be swapped, e.g. color \textit{blue} may be changed to \textit{green}. For example,  when $\textit{p}=0.2$, the probability of having the whole instruction correct is  $0.8^{4}\approx 0.4$. The resulting agent performance is depicted in \autoref{fig:noisyher} (left).
 
 The agent performs 80\% as well as an agent with perfect expert feedback even when the mapping function has a 50\% noise-ratio per attribute. Surprisingly, even highly noisy mappers, with a 80\% noise-ratio, still provides an improvement over vanilla DQN-agents.
 Hence, HER can be applied even when relabelling trajectories with partially correct goals.

 We also examine whether this robustness may be induced by the environment properties (e.g. attribute redundancy) rather than HER. We thus compute the number of discriminative features required to pick the correct object. On average, an object can be discriminated with 1.7 features in our setting - which eases the training, but any object shares at least one property with any other object 70\% of the time - which tangles the training. Besides, the agent does not know which features are noisy or important. Thus, the agent still has to disentangle the instructions across trajectories in the replay buffer, and this process is still relatively robust to noise. 
 

\vspace{-2em}
\begin{figure*}
     \centering
     \begin{minipage}{0.48\textwidth}
         \centering
         \includegraphics[width=\textwidth]{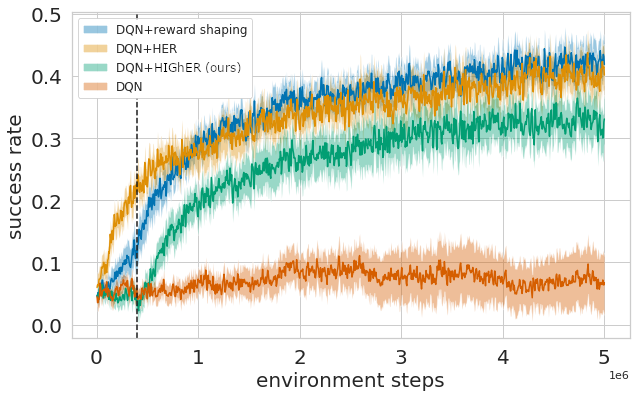}
     \end{minipage}
     \hfill
     \begin{minipage}[h]{0.48\textwidth}
         \centering
         \includegraphics[width=\textwidth]{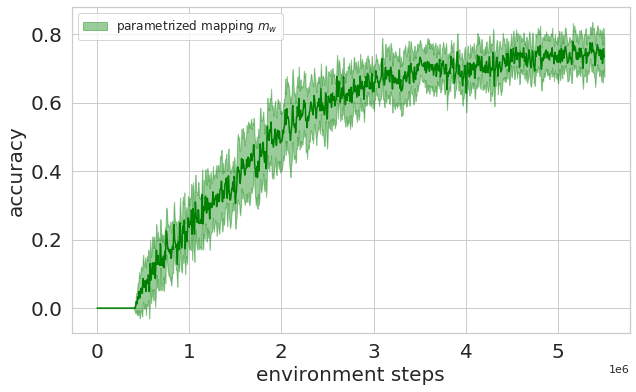}
     \end{minipage}
        \caption{\textbf{Left}: learning curves for DQN, DQN+HER, DQN+\him in a 10x10 gridworld with 10 objects with 4 attributes. The instruction generator is used after the  vertical bar. \textbf{Right}: the mapping accuracy for the prediction of instructions. $m_{\vw}$ starts being trained after collecting 1000 positive trajectories.
        Results are averaged over 5 seeds with one standard deviation.}
        \label{fig:text_her}
\vspace{-1.5em}
\end{figure*}

\vspace{2em}
 \subsubsection{Learning an instruction generator}\label{subsec:ig}

  We briefly analyze the sample complexity and generalization properties of the instruction generator. If training the mapping function is more straightforward than learning the agent policy, then we can thus use it to speed up the navigation training. 
 
 We first split the set of missions $G$ into two disjoint sets $G_{train}$ and $G_{test}$. Although all object features are present in both sets, they contain dissimilar combinations of target objects. For instance, \textit{blue}, \textit{dark}, \textit{key}, and \textit{large} are individually present in instructions of $G_{train}$ and $G_{test}$ but the instruction to get a \textit{large dark blue key} is only in $G_{test}$. We therefore assess whether a basic compositionality is learned. In the following, we use train/split ratio of 80/20, i.e., 240 vs 60 goals.
 
  
  We here observe than 1000 positive episodes are necessary to reach around 20\% accuracy with our model, and 5000 pairs are enough to reach 70\% accuracy.  
 The instruction generator also correctly predicts unseen instructions even with fewer than 1000 samples and the accuracy gap between seen and unseen instructions slowly decrease during training, showing basic compositionality acquisition. 
 As further discussed in \autoref{sec:related}, we here use a vanilla mapping architecture to assess the generality of our \him, and more advanced architectures may drastically improve sample complexity~\cite{bahdanau2018systematic}.

\subsection{\him for instruction following}

 In the previous section, we observe  that: (1) HER is robust to noisy relabeled goals, (2) an instructor generator requires few positive samples to learn basic language compositionality. We thus here combine those two properties to execute \him, i.e. jointly learning the agent policy and language prediction in a online fashion for instruction following.

\paragraph{Baselines}\label{subsec:mcontrol}
We want to assess if the agent benefits from learning an instruction generator and using it to substitute goals as done in HER. We denote this approach DQN+\him. We compare our approach to DQN without goal substitution (called DQN) and DQN with goal substitution from a perfect mapping provided by an external expert (called DQN+HER) available in the BabyAI environment. We emphasize again that it is impossible to have an external expert to apply HER in the general case. Therefore, DQN is a lower bound that we expect to outperform, whereas DQN+HER is the upper bound as the learned mapping can not outperform the expert. Note that we only start using the parametrized mapping function after collecting 1000 positive trajectories, which is around 18\% validation accuracy. Finally, we compute an additional DQN baseline denoted DQN+reward: we reward the agent with 0.25 for each matching properties when picking a object given an instruction. It enforces a hand-crafted curriculum and dramatically reduces the reward sparsity, which gives a different perspective on the current task difficulty.

\paragraph{Results}\label{subsec:results}
In \autoref{fig:text_her} (left), we show the success rate of the benchmarked algorithms per environment steps. We first observe that DQN does not manage to learn a good policy, and its performance remains close to that of a random policy. On the other side, DQN+HER and DQN+reward quickly manage to pick the correct object 40\% of the time. Finally, DQN+\him sees its success rates increasing as soon as we use the mapping function, to rapidly perform nearly as well as DQN+HER. \autoref{fig:text_her} (right) shows the performance accuracy of the mapping generator by environment steps. We observe a steady improvement of the accuracy during training before reaching 78\% accuracy after 5M steps. In the end, DQN+\him outperforms DQN by using the exact same amount of information, and even matches the conceptual upper bond computed by DQN+HER. Besides, \him does not alter the optimal policy which can occur when reshaping the reward~\cite{ng1999policy}. As stated in \autoref{subsec:mcontrol}, a gap remains between DQN+HER and HIGhER as the latter is building an approximate model of the instruction, thus sometimes failing to relabel correctly as pointed out in \autoref{fig:noisyher}

\subsection{Discussion}\label{subsec:discussion}

\paragraph{Improvements over DQN} As observed in the previous noisy-HER experiment, the policy success rate starts increasing even when the mapping accuracy is 20\%, and DQN+\him becomes nearly as good as DQN+HER despite having a maximum mapping accuracy of 78\%. It demonstrates that DQN+\him manages to trigger the policy learning by better leveraging environment signals compared to DQN. 
As the instruction generator focuses solely on grounding language, it quickly provides additional training signal to the agent, initiating the navigation learning process. 

%

\paragraph{Generator Analysis} We observe that the number of positive trajectories needed to learn a non-random mapping $m_{\vw}$ is lower than the number of positive trajectories needed to obtain a valid policy with DQN (even after 5M environment steps the policy has 10\% success rate). Noticeably, we artificially generate a dataset in \autoref{subsec:ig} to train the instruction generator, whereas we follow the agent policy to collect the dataset, which is a more realistic setting. 
For instance, as the instructor generator is trained on a moving dataset, it could overfit to the first positive samples, but in practice it escapes from local minima and obtains a final high accuracy. 

Different factors may also explain the learning speed discrepancy: supervised learning has less variance than reinforcement learning as it has no long-term dependency. The agent instructor generator can also rely on simpler neural architectures than the agent. Although \him thus takes advantage of those training facilities to reward the agent ultimately.

\paragraph{Robustness} Finally, we observe a virtuous circle that arises. As soon as the mapping is correct, the agent success rate increases, initiating the synergy. The agent then provides additional ground-truth mapping pairs, which increases the mapping accuracy, which improves the quality of substitute goals, which increases the agent success rate further more. As a result, there is a natural synergy that occurs between language grounding and navigation policy as each module iteratively provides better training samples to the other model.
If we ignore time-out trajectories, around 90\% of the trajectories are negative at the beginning of the training. As soon as we start using the instruction generator, 40\% the transitions are relabelled by the instructor generator, and  10\% of the transitions belong to positive trajectories. As training goes, this ratio is slowly inverted, and after 5M steps, there is only 15\% relabelled trajectories left while 60\% are actual positive trajectories.

\begin{figure}[ht]
 \begin{center}
 \includegraphics[width=0.7\linewidth]{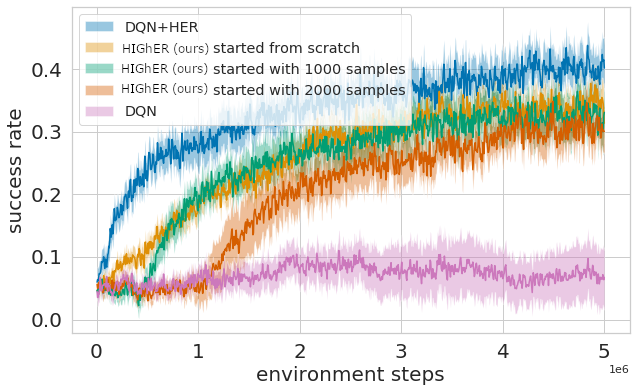}
 \end{center}
 \caption{The instruction generator is triggered after collecting 0, 1000 and 2000 positive trajectories (i.e, approximately 0\%, 20\%, 50\% accuracy). Even when the instruction generator is not accurate, the policy still makes steady progress and the final success rate is not impacted. Delaying the generator instructor does not provide additional benefit}
 \label{fig:theracc}
\end{figure}
\begin{figure}[ht]
    \vspace{-1em}
    \centering
    \includegraphics[width=0.8\columnwidth]{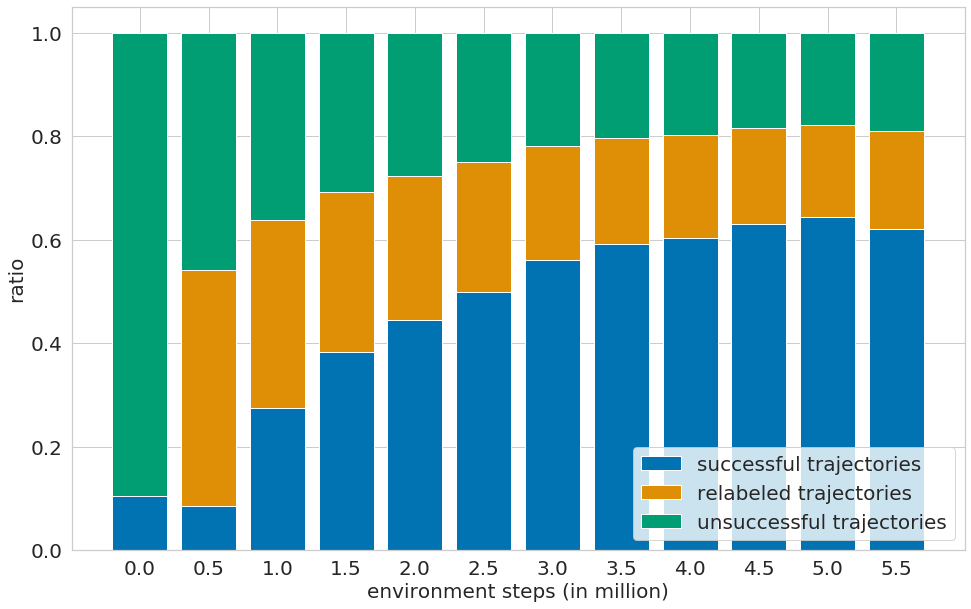}
        \caption{Transition distributions in the replay buffer between successful, unsuccessful and relabeled trajectories. We remove time-out trajectories for clarity, which accounts for $54\%$ of the transition in average ($\pm3\%$ over training).}
\vspace{-1.5em}
\end{figure}

\vspace{-0.3em}
\subsection{Limitations}

Albeit generic, \him also faces some inherent limitations. 
From a linguistic perspective, \him cannot transcribe negative instructions (\emph{Do not pick the red ball}), or alternatives (\emph{Pick the red ball or the blue key}) in its current form. However, this problem could be alleviated by batching several trajectories with the same goal. Therefore, the model would potentially learn to factorize trajectories into a single language objective. 
On the policy side, \him still requires a few trajectories to work, and it thus relies on the navigation policy. In other word, historical HER could be applied in the absence of reward signals, while \him only alleviate the sparse reward problem by better leveraging successful trajectories. A natural improvement would be to couple \him with other exploration methods, e.g, intrinsic motivation~\cite{bellemare2016unifying} or DQN with human demonstration~\cite{hester2018deep}. 
Finally, under-trained goal generators might hurt the training in some environments although we did not observe it in our setting as shown in \autoref{fig:theracc}. 
However, a simple validation accuracy allows to circumvent this risk while activating the goal mapper (More details in \autoref{algo:him}). We emphasize again that the instruction generator can be triggered anytime to kick-start the learning as the it is independent of the agent. 

\vspace{-1em}
\section{Related Work}\label{sec:related}
\vspace{-0.5em}

Instruction following have recently drawn a lot of attention following the emergence of several 2D and 3D environments~\cite{babyai_iclr19,brodeur2017home,anderson2018vision}. This section first provides an overview of the different approaches, i.e, fully-supervised agent, reward shaping, auxiliary losses, before exploring approaches related to \him.

\vspace{-0.3em}
\subsection{Vision and Language Navigation} Instruction following is sometimes coined as \textit{Vision and Language Navigation} tasks in computer vision~\cite{anderson2018vision,wang2019reinforced}.
Most strategies are based on imitation learning, relying on expert demonstrations and knowledge from the environment. For example, \cite{zang2018translating} relate instructions to an environment graph, requiring both demonstrations and high-level navigation information. 
Closer to our work, \cite{fried2018speaker} also learns a navigation model and an instruction generator, but the latter is used to generate additional training data for the agent. The setup is hence fully supervised, and requires human demonstrations. These policies are  sometimes finetuned to improve navigation abilities in unknown environments. Noticeably, \cite{wang2019reinforced} optimizes their agent to find the shortest path by leveraging language information. The agent learns an instruction generator, and they derive an intrinsic reward by aligning the generator predictions over the ground truth instructions. Those approaches complete long sequences of instructions in visually rich environments but they require a substantial amount of annotated data. In this paper, we intend to discard human supervision to explore learning synergies. Besides, we needed a synthetic environments with experts to evaluate \him. Yet, \him could be studied on natural and visually rich settings by warm-starting the instruction generator, and those two papers give a hint that \him could scale up to larger environment. Recent works using pretrained language model \cite{devlin2019bert, hill2020human} could also complement HIGhER both in the generator and in the instruction understanding.

\vspace{-0.3em}
\subsection{IRL for instruction following}
\cite{bahdanau2019learning} learn a mapping from $<$instruction,~state$>$ to a reward function. The method's aim is to substitute the environment's reward function when instructions can be satisfied by a great diversity of states, making hand-designing reward function tedious. Similarly, \cite{fu2019from} directly learn a reward function and assess its transferability to new environments. Those methods are complementary to ours as they seek to transfer reward function to new environment and we are interested in reducing sample complexity.

\vspace{-0.3em}
\subsection{Improving language compositionality} \him heavily relies on leveraging the language structure in the instruction mapper toward initiating the learning synergy. For instance, \cite{bahdanau2018systematic} explore the generalization abilities of various neural architectures. They show that the sample efficiency of feature concatenation can be considerably improved by using feature-wise modulation~\cite{perez2018film}, neural module networks~\cite{andreas2016neural} or compositional attention networks~\cite{arad2018compositional}. In this spirit, \cite{bahdanau2019learning} take advantage of these architectures to quickly learn a dense reward model from a few human demonstrations in the instruction following setup. Differently, the instructor generator can also be fused with the agent model to act as an auxiliary loss, reducing further the sparse reward issue.

\vspace{-0.3em}
\subsection{HER variants} HER has been extended to multiple settings since the original paper. These extensions deal with automatic curriculum learning~\cite{liu2018competitive}, dynamic goals~\cite{fang2019dher}, or they adapt goal relabelling to policy gradient methods~\cite{rauber2018hindsight}. Closer to our work, \cite{sahni2019visual} train a generative adversarial network to hallucinate visual near-goals state over failed trajectories. However, their method requires heavy engineering as visual goals are extremely complex to generate, and they lack the compact generalization opportunities inherent to language. \cite{chan2019actrce} also studies HER in the language setting, but the authors only consider the context where a language expert is available. 

\vspace{-0.2em}
\subsection{Conditioned Language Policy} 
There have been other attempts to leverage language instruction to improve the agent policy. For instance, \cite{jiang2019language} computes a high-level language policy to give textual instruction to a low-level policy, enforcing a hierarchical learning training. The authors manage to resolve complicated manipulating task by decomposing the action with language operation. The language mapper performs instruction retrieval into a predefined set of textual goals and yet, the low-level policy benefits from language compositionality and is able to generalize to unseen instructions, as mentioned by the authors. \cite{co2018guiding} train an agent to refine its policy by collecting language corrections over multiple trajectories on the same task. While the authors focus their effort on integrating language cues, it could be promising to learn the correction function in a \him fashion.

\section{Conclusion}
We introduce \himlong (\him) as an extension to HER for language. We define a protocol to learn a mapping function to relabel unsuccessful trajectories with predicted consistent language instructions. 
We show that \him nearly matches HER performances despite only relying on signals from the environment. We provide empirical evidence that \him manages to alleviate the instruction following task by jointly learning language grounding and navigation policy with training synergies. \him has mild underlying assumptions, and it does not require human data, making it valuable to complement to other instruction following methods. More generally, \him can be extended to any goal modalities, and we expect similar procedures to emerge in other setting.

\section*{Acknowledgements}

The authors would like to acknowledge the stimulating research environment of the SequeL INRIA Project-Team. Special thanks to Edouard Leurent, Piotr Mirowski and Florent Altché for fruitful discussions and reviewing the final manuscript.

We acknowledge the following agencies for research funding and computing support: Project BabyRobot (H2020-ICT-24-2015, grant agreement no.687831), and CPER Nord-Pas de Calais/FEDER DATA Advanced data science and technologies 2015-2020.

Experiments presented in this paper were carried out using the Grid'5000 testbed, supported by a scientific interest group hosted by Inria and including CNRS, RENATER and several Universities as well as other organizations (see https://www.grid5000.fr).

\printbibliography

\end{document}

%% file: math_commands.tex
\usepackage{amsmath}
\usepackage{amsfonts}
\usepackage{bm}

\def\eqref#1{equation~\ref{#1}}

\def\1{\bm{1}}

\def\vw{{\bm{w}}}

\DeclareMathAlphabet{\mathsfit}{\encodingdefault}{\sfdefault}{m}{sl}
\SetMathAlphabet{\mathsfit}{bold}{\encodingdefault}{\sfdefault}{bx}{n}

\def\gA{{\mathcal{A}}}

\def\gG{{\mathcal{G}}}

\def\gS{{\mathcal{S}}}

\def\sA{{\mathbb{A}}}

\newcommand{\R}{\mathbb{R}}

